\documentclass[conference]{IEEEtran}
\IEEEoverridecommandlockouts
\usepackage{cite}
\usepackage{amsmath,amssymb,amsfonts}
\usepackage{algorithmic}
\usepackage{graphicx}
\usepackage{textcomp}
\usepackage{xcolor}
\usepackage{hyperref}
\usepackage{array}
\usepackage{stfloats}
\usepackage{url}
\usepackage{verbatim}
\usepackage{xstring}
\usepackage{subfigure}
\usepackage{epsfig} 
\usepackage{times} 
\usepackage{mathtools}
\usepackage{mwe}
\usepackage{float}
\usepackage{adjustbox}

\def\BibTeX{{\rm B\kern-.05em{\sc i\kern-.025em b}\kern-.08em
    T\kern-.1667em\lower.7ex\hbox{E}\kern-.125emX}}
\begin{document}

\title{A Resilient Navigation and Path Planning System for High-speed Autonomous Race Cars \\
\thanks{\textsuperscript{1} School of Electrical Engineering, Korea Advanced Institute of Science and Technologies (KAIST), Daejeon, Republic of Korea
        {\texttt{\{lee.dk, cy.jung, hynkis, hcshim\}@kaist.ac.kr}}}
\thanks{\textsuperscript{2} Robotics Program, KAIST, Daejeon, Republic of Korea
        {\texttt{finazzi@kaist.ac.kr}}
        }
\thanks{\textsuperscript{*} Corresponding Author}
\thanks{This work is partially supported by the Institute of Information \& Communications Technology Planning Evaluation (IITP) grant funded by the Korean government (MSIT, 2021-0-00029).}
}

\author{\IEEEauthorblockN{Daegyu Lee\textsuperscript{1}, Chanyoung Jung\textsuperscript{1}, Andrea Finazzi\textsuperscript{2}, Hyunki Seong\textsuperscript{1}, and D. Hyunchul Shim\textsuperscript{1*}}
}

\maketitle

\begin{abstract}
This paper describes a resilient navigation and planning system used in the Indy Autonomous Challenge (IAC) competition.
The IAC is a competition where full-scale race cars run autonomously on Indianapolis Motor Speedway(IMS) up to 290 km/h (180 mph). 
Race cars will experience severe vibrations. Especially at high speeds. 
These vibrations can degrade standard localization algorithms based on precision GPS-aided inertial measurement units.
Degraded localization can lead to serious problems, including collisions.
Therefore, we propose a resilient navigation system that enables a race car to stay within the track in the event of localization failures.
Our navigation system uses a multi-sensor fusion-based Kalman filter.
We detect degradation of the navigation solution using probabilistic approaches to computing optimal measurement values for the correction step of our Kalman filter.
In addition, an optimal path planning algorithm for obstacle avoidance is proposed. 
In this challenge, the track has static obstacles on the track. 
The vehicle is required to avoid them with minimal time loss.
By taking the original optimal racing line, obstacles, and vehicle dynamics into account, we propose a road-graph-based path planning algorithm to ensure that our race car can perform efficient obstacle avoidance.
The proposed localization system was successfully validated to show its capability to prevent localization failures in the event of faulty GPS measurements during the historic world's first autonomous racing at IMS. 
Owing to our robust navigation and planning algorithm, we were able to finish the race as one of the top four teams while the remaining five teams failed to finish due to collisions or out-of-track violations. 
\end{abstract}


\section{Introduction}
The Indy autonomous challenge (IAC) was held on October 23, 2021. In this competition, race cars were modified so that they could be driven autonomously. 
Technologies were developed for this event that enabled a full-scale race car to complete laps on a racetrack autonomously. These technologies push the current state of autonomous driving to new heights.
Each team received the same modified Dallara IL-15, called the AV-21. 
Teams were required to develop autonomous driving software for the AV-21, which is capable of reaching 290 $kph$ (180 $mph$).
\begin{figure}[t]
    \centering
    \includegraphics[width=1.0\columnwidth]{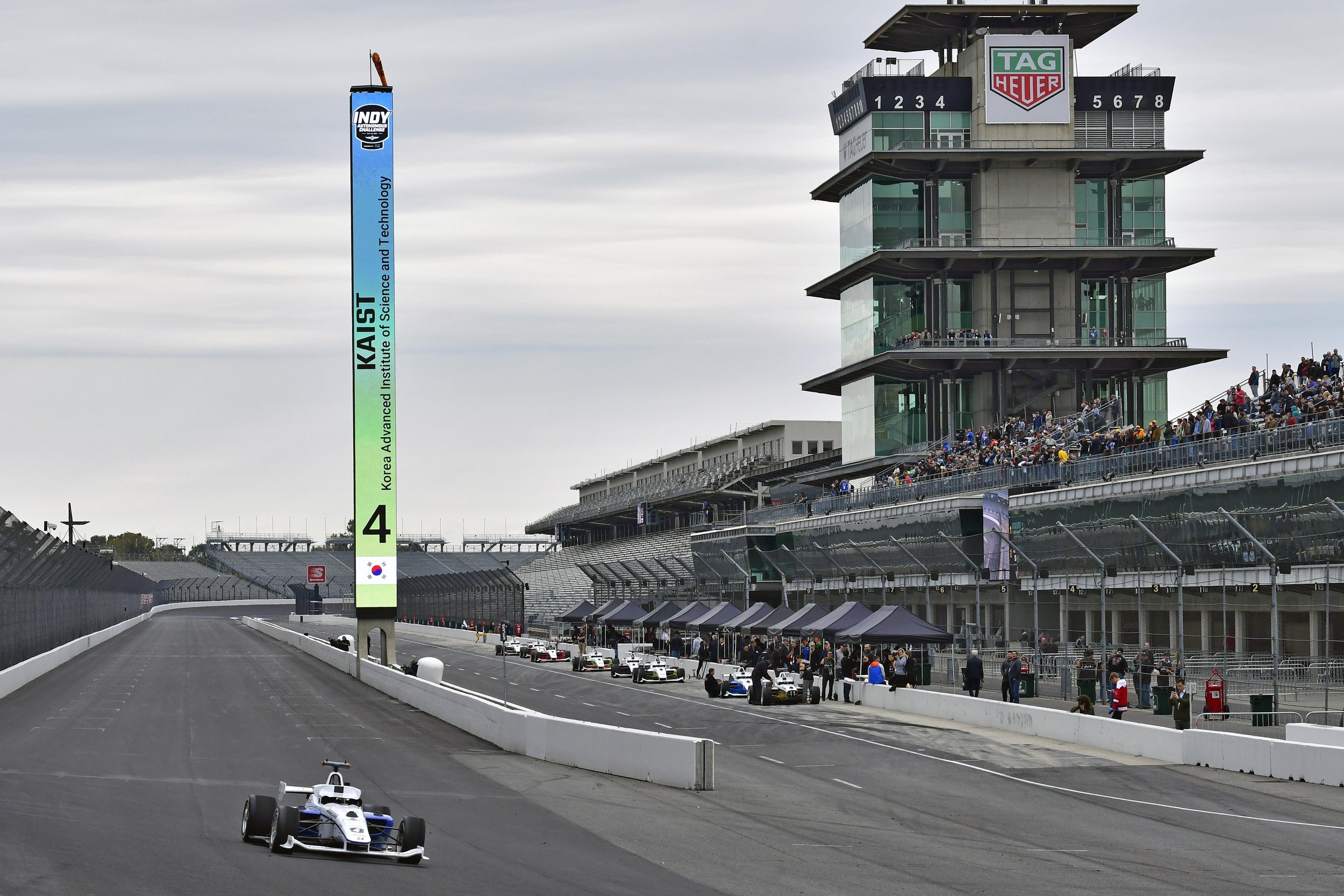}
    \caption[Dallara AV-21 of Team KAIST at the Indianapolis Motor Speedway (IMS).]{Dallara AV-21 of Team KAIST at the Indianapolis Motor Speedway (IMS).}
    \label{fig:main}
\end{figure}
\\
In this race, the GPS sensor is the most essential sensor for localization. 
Therefore, the teams had to address the sensor noise issue which degraded localization performance.
Moreover, the drift caused by strong vibrations during high-speed driving can lead to serious accidents.
For example, when a 120-$inch$ (3.048 $m$) wheel-base race car is running at 180 $kph$ (50.0 $mps$), lateral control with 3.0 $\deg$ results in 22.2 $m$ lateral transition.
This lateral deviation error caused by GPS degradation can lead to critical crashes when taking into account the IMS track width of 15 $m$ and 18 $m$ respectively.
Indeed there were a number of cars that collided with walls or seriously diverted from the track reportedly due to GPS error.
\\
To overcome this, a reliable localization system using a multi-sensor fusion Kalman filter algorithm is proposed. The algorithm combines two GPS units and LiDAR sensors to resolve localization problems.
We identify the GPS reliability in real-time through a probabilistic Bayesian approach and then implement a stable localization algorithm.
We also implement a wall-following navigation algorithm as extended resilient navigation when the GPS divergences.
The wall detection-based resilient navigation system was operational when the GPS sensor was temporarily disabled during the competition.
Furthermore, we also present our road-graph-searching-based obstacle avoidance algorithm to guarantee that the vehicle runs within in-bounded conditions.
\\
The remainder of this paper is organized as follows. Section \ref{sec:related} introduces related IAC studies. 
Section \ref{sec:methods} describes our high-speed, resilient localization system and motion planning algorithm for obstacle avoidance.
The experimental results, including track-time for test and competition, are discussed in Section \ref{sec:results}. 

\section{Related works}
\label{sec:related}
Autonomous racing is an emerging field of high-performance robotics that can push the boundaries of autonomous systems technologies.
In this section, we introduce recent autonomous racing studies for the edge cases of autonomous driving. \\
A comprehensive overview of the current autonomous racing platforms, emphasizing the software-hardware co-evolution to the current stage, is presented in \cite{betz2022autonomous}.
After IAC 2021 and CES 2022, each team presented their full stack of autonomous systems and approaches in \cite{betz2022tum, spisak2022robust}. 
TUM Autonomous Motorsport team, winner of the 2021 IAC, also shared its research on optimal control systems \cite{wischnewski2022tube, wischnewski2022indy} and planning \cite{heilmeier2019minimum, christ2021time, hermansdorfer2019concept, herrmann2020minimum}.
\\
In addition, a behavior planning system will be one of the most important systems for overtaking or making decisions in autonomous racing when it becomes analogous to human-driven racing.
Therefore, we examined the game-based predictor to predict racing competitors' future trajectories \cite{jung2021game}.


\section{Methods}
\label{sec:methods}
\subsection{Multi-sensor fusion Kalman filter}
\label{sec:kalman}
\begin{figure}[t]
    \centering
    \includegraphics[width=0.75\columnwidth]{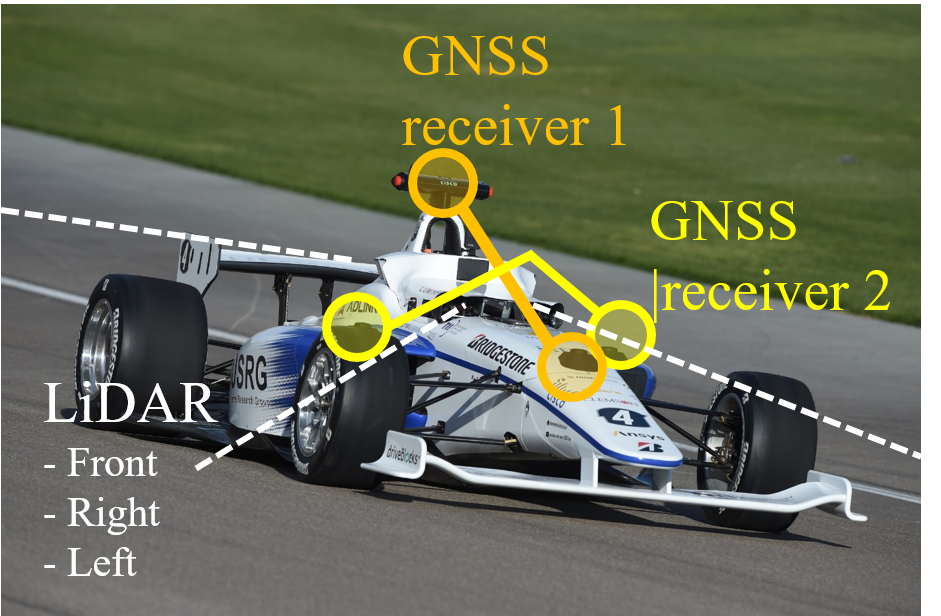}
    \caption[Race car sensor configuration of race car. The positions of the two GPS units and three LiDARs are illustrated.]{Sensor configuration of race car. The positions of two GPS units and three LiDARs are illustrated.}
    \label{fig:sensors_with_race_car}
\end{figure}

The race car is equipped with two RTK GPS, GPS-aided inertial measurement unit (IMU) and three LiDARs, as shown in Fig. \ref{fig:sensors_with_race_car}.
We propose our multi-sensor fusion Kalman filter algorithm to compensate for GPS measurement noise caused by strong vibrations.

\subsubsection{Problem definition}
The Kalman filtering process of time-varying stochastic control system with $k$ multiple measurements is given by
\begin{equation}
\begin{aligned}
    & \mathbf{x}_{t+1} = \mathbf{F}_{t}\mathbf{x}_{t} + \mathbf{B}_{t}\mathbf{u}_{t} + \mathbf{\omega}_{t}, \\
    & \mathbf{y}_{t}^{k} = \mathbf{H}_{t}^{k}\mathbf{x}_{t} + \mathbf{v}_{t}, 
\end{aligned}
\label{eq_kalman_basic}
\end{equation}
where $\mathbf{x}_{t} \in \mathbb{R}^n$ is the state, $\mathbf{y}_{t}^{k} \in \mathbb{R}^m$ is the $k$-th measurement, and $\mathbf{u}_{t} \in \mathbb{R}^l$ is the control input. $\mathbf{\omega}_{t}$ and $\mathbf{v}_{t}$ are white noise. 
We can address our localization problem as the posterior probability in terms of the prior likelihood \cite{charles2017kalman}.
Our problem definition starts with optimizing the overall state of $\hat{\mathbf{x}}_t$ with multiple measurement $\mathbf{y}_{t}^{k}$ : 
\begin{equation}
\begin{aligned}
    & \{\hat{\mathbf{x}}_t\}_{t \in [1,t]} = argmax[\prod_{i=0}^{t} p(\mathbf{x}_{i}|\mathbf{x}_{i-1})p(\mathbf{y}_{i}^{k}|\mathbf{x}_{i})p(\mathbf{y}_{0}^{k}|\mathbf{x}_{0})p(\mathbf{x}_{0})].
\end{aligned}
\label{eq_kalman_1}
\end{equation}
We define our localization problem as finding an optimal solution at the current time $t$; a marginalization can be represented as follows:
\begin{equation}
\begin{aligned}
    \hat{\mathbf{x}}_t = & \arg\max_{\mathbf{x}_{t}}[\int p(\mathbf{y}_{i}^{k}|\mathbf{x}_{t})p(\mathbf{x}_{t}|\mathbf{x}_{t-1}) \\ 
    & \prod_{i=0}^{t} p(\mathbf{x}_{i}|\mathbf{x}_{i-1})p(\mathbf{y}_{i}^{k}|\mathbf{x}_{i})p(\mathbf{y}_{0}^{k}|\mathbf{x}_{0})p(\mathbf{x}_{0})d\{\mathbf{x}_{l}\}_{l \in [1,t-1]}],
\end{aligned}
\label{eq_kalman_2}
\end{equation}
where we assume each variable as independent and identically distributed.
We further simplify Eq. \ref{eq_kalman_2} as
\begin{equation}
\begin{aligned}
    \hat{\mathbf{x}}_t \propto & \arg\max_{\mathbf{x}_{i}}[\int p(\mathbf{y}_{i}^{k}|\mathbf{x}_{i})p(\mathbf{x}_{t}|\mathbf{x}_{t-1})d\{\mathbf{x}_{l}\}_{l \in [1,t-1]}].
\end{aligned}
\label{eq_kalman_3}
\end{equation}
Using Bayes' theorem, we propose a hyper-parameter $\theta$ for localization to conditionally update the measurement as follows: 
\begin{equation}
\begin{aligned}
    \hat{\mathbf{x}}_t \propto & \arg\max_{\theta}[\int p(\mathbf{y}_{i}^{k}|\theta^{k})p(\theta^{k}|\mathbf{x}_{i})p(\mathbf{x}_{t}|\mathbf{x}_{t-1})d\{\mathbf{x}_{l}\}_{l \in [1,t-1]}].
\end{aligned}
\label{eq_kalman_4}
\end{equation}
Intuitively, Eq. \ref{eq_kalman_4} is the likelihood of the $k$-th measurement $\mathbf{y}_{t}^{k}$ over all possible parameter values, weighted by the prior $p(\theta^{k}|\mathbf{x}_{i})$.
If all localization hyper-parameters $p(\theta^{k})$ assign high probability to the $k$-th measurement data, then this is a reasonable criteria for multiple measurements.
We define our multi-sensor fusion algorithm for high-speed localization as finding $p(\mathbf{y}_{i}^{k}|\theta^{k})$ and $p(\theta^{k}|\mathbf{x}_{i})$.
Therefore, we avoid Kalman filter updates based on degraded measurements by determining a poor signal quality computing hyper-parameter $\theta$.

\subsubsection{Bayesian update selection}
We compute $p(\theta^{k}|\mathbf{x}_{t})$ using the Mahalanobis distance $\Delta_{k}$ \cite{de2000mahalanobis} as follows:
\begin{equation}
\begin{aligned}
    p(\theta^{k}|\mathbf{x}_{t})  
    & \triangleq - (\mathbf{x}_{t-1} - \mathbf{z}^{k})^T\Sigma^{-1}(\mathbf{x}_{t-1} - \mathbf{z}^{k}), \\
    & = - \Delta_{k}
\end{aligned}
\label{eq_mahala}
\end{equation}
where $\mathbf{z}^{k}$ is the $k$-th measurement and $\Sigma^{-1} = \Lambda$ is the precision matrix.
We interpret $\Delta_{k}$ as a Euclidean distance in an orthogonally transformed new coordinate frame to identify whether $\mathbf{z}^{k}$ is a reasonable input or not. 
\\
In addition, let $p(\mathbf{y}_{i}^{k}|\theta^{k}, \mathbf{x}_{i})$ be defined conditional to $\Delta_{k}$ as follows:
\begin{equation}
    \begin{aligned}
        p(\mathbf{y}_{i}^{k}|\theta^{k}, \mathbf{x}_{i})  
        = 
        \begin{cases}
        &  \mathbf{z}^0 \hfill \text{ if } \forall\Delta_{k} \leq \epsilon \\
        &  \lambda_{i} \cdot \mathbf{z}^i \hfill \text{ if } \forall\Delta_{k} > \epsilon \text{ and } \forall\Delta_{k} \leq \delta \\
        &  \mathbf{z}^{k} \hfill \text{ if } \Delta_{k} \leq \delta \text{ and } \forall\Delta_{ \sim k} > \delta \\
        & \lambda_{reject} \hfill \text{ if } \forall\Delta_{k} > \delta, \\    
        \end{cases}
    \end{aligned}
    \label{eq_conditional_measure}
\end{equation}
where $\lambda_{k} = 1 - \mathbf{z}^k / \sum(\mathbf{z}^i)$, $\delta$, and $\epsilon$ are the measurement update weight, hyper-parameter to determine qualified measurement, and error($\epsilon << \delta$), respectively.
If all  $\Delta_{k}$ is less than $\epsilon$, we use one of the measurements because it is already qualified.
If all $\Delta_{k}$ is less than $\delta$, we implement the weighted sum of measurement, assuming that all the measurement is in reasonable condition.
In contrast, if $\Delta_{k}$ is less than $\delta$, but $\Delta_{\sim k}$ is greater than $\delta$, we use the only feasible measurement $z^k$.
Lastly, if all the distances are unsuitable for measurement updates, a warning alarm is raised to be ready for the resilient localization system. This is discussed in the next section \ref{sec:sub_wall_detection}.
If the new measurement status is rejected, we update our state only depending on control input $\mathbf{u}_t$. $\mathbf{z}^k$ is not utilized in this condition.

\subsection{Wall detection for resilient navigation}
\label{sec:sub_wall_detection}
When our proposed multi-sensor fusion Kalman filter continually computes $p(\mathbf{y}_{i}^{k}|\theta^{k}, \mathbf{x}_{i})$ as $\lambda_{reject}$, we consider this status as a  positioning degraded situation.
To deal with these critical situations, we designed our race car such that it follows along the racetrack wall. This allows the vehicle to avoid colliding with the wall.
To extract walls in a racetrack, we propose a wall detection algorithm comprising of ground removal and wall clustering algorithms. The algorithm helps navigate the wall resiliently when a degraded measurement is detected.

\subsubsection{Vertical feature extraction for ground filtering}
\begin{figure*}[!t]
    \centering
    \subfigure[]{
        \includegraphics[width=0.53\columnwidth]{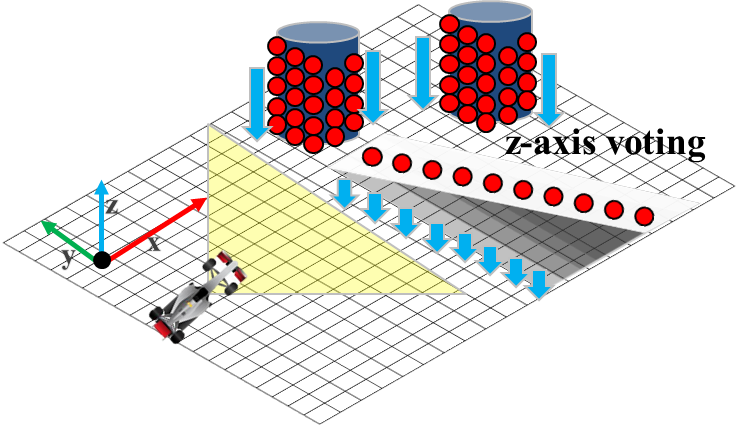}
        \label{fig:z_vote}
    }
    \subfigure[]{
        \includegraphics[width=0.53\columnwidth]{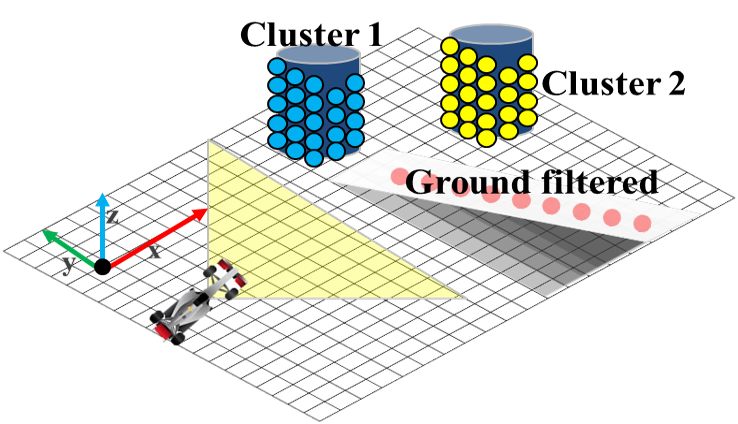}
        \label{fig:ground_filter}
    }
    \subfigure[]{
        \includegraphics[width=0.53\columnwidth]{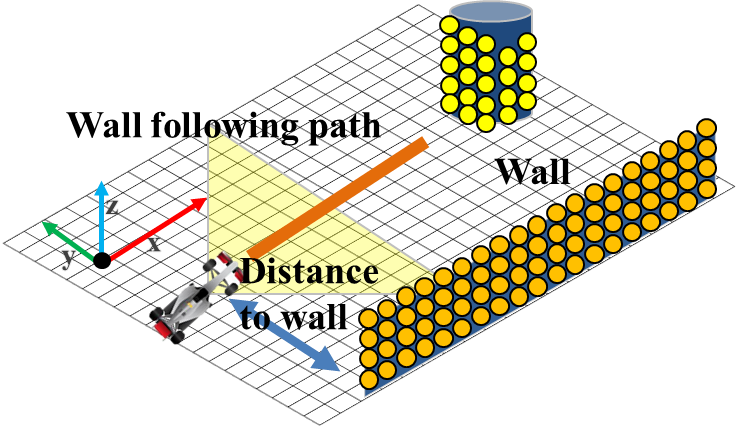}
        \label{fig:wall_detection}
    }    
    \label{fig:point}
    \caption[Description of process for ground points filtering, clustering, and wall detection]{
    Description of process for ground points filtering, clustering, and wall detection.
    (a) Ground filtering based on a z-axis voting algorithm. The number of voting to grid-cell is compared to a hyper-parameter to extract the vertical features. In the illustration, points corresponding to the banked road have a few counts.  
    (b) Ground-filtered points are illustrated. In addition, the Euclidean-distance clustering algorithm is implemented.   
    (c) The length of clusters is used, assuming that the longest right-side cluster is the wall. Points of the wall are computed to estimate the wall curvature and distance to the wall.
    }
\end{figure*}

Generally racetrack roads are banked to enable faster corner speeds.
Thus, the ground filtering algorithm for LiDAR points has to consider this road gradient. \\
We propose a novel vertical feature extractor using a hashing algorithm to consider the sparsity of LiDAR point-cloud data. 
Let $B \subset \mathbb{R}^3$ define the vehicle body coordinates.
Here, $B$ annotated values indicate the information obtained from the vehicle body’s origin---i.e., the center point of the rear axle. 
We also define the voxel-filtered LiDAR points $\mathbf{p}^B_{t}= \{p_1^B, \dots, p_k^B\}$ at time $t$, where $p_i^B$ is a voxelized point from the incoming LiDAR points.
After voxelization, we project $\mathbf{p}^B_{t}$ to the 2-D grid to vote the points corresponding to the grid-cell, as shown in Fig. \ref{fig:z_vote}.
Moreover, we propose a hashing algorithm during voting to account for the sparsity of the point cloud.
\begin{equation}
\begin{aligned}
    & H(p_i^B) = index, \\
    & H(index) = p_i^B,
\end{aligned}
\label{eq_hash_table}
\end{equation}
where hash function $H(x)$ maps the value $x$ at the table.
We iterate voxelized points $\mathbf{p}^B_{t}$ to vote on the corresponding grid-cell using $f(p_i^B) = g(u_i, v_i, \mathbf{n}_{index})$ where $\mathbf{n}_{index}$ contains the number of points voted and its hashing-index.
Thus, by comparing $\mathbf{n}_{index}$ size with hyper-parameter, we can efficiently filter out ground points $\mathbf{p}^{ground}_{t}$ from $\mathbf{p}^B_{t}$ in real-time without matrix computation for plane extraction.
\subsubsection{Wall following navigation}
To extract the plane feature of the wall from the point-cloud, we implement a random sample consensus (RANSAC)-variant algorithm \cite{qian2014ncc, li2017improved}.
However, a distance parameter determines the inlier or outlier points, which is unsuitable for curved wall areas.
Our approach uses a common tree-based Euclidean distance algorithm to find a wall from the ground-filtered points $\mathbf{p}^{filtered}_{t}$, as shown in Fig. \ref{fig:wall_detection}.
We implement a CUDA-based Euclidean distance clustering algorithm \cite{karbhari2018gpu, nguyen2020fast} to find cluster $C_j$ \cite{rusu2010semantic}:
\begin{equation}
\begin{aligned}
    & C_j = \arg \min_{i}\parallel{p}^{filtered}_i - \mu_j\parallel_{2}, \\
    & \mu_{j} = \frac{\sum_{i=1}^{m}1\{c^i=j\}{p}^{filtered}_i}{\sum_{i=1}^{m}1\{c^i=j\}},
\end{aligned}
\label{eq_clustering}
\end{equation}
where Eq. \ref{eq_clustering} is the subject condition for clustering algorithm.
We assume the longest cluster as wall cluster $\mathbf{w}_i(x,y)$ :
\begin{equation}
\begin{aligned}
    \mathbf{w}_i(x,y) = \mathbf{H}_z \cdot \arg\max_{x}(h(C_{j}(x_i,y_i,z_i))),
\end{aligned}
\label{eq_wall_extract}
\end{equation}
where $h(x)$ extracts the length of clusters, and $\mathbf{H}_z$ is the z-directional projection matrix. \\
We then obtain a coefficient of the polynomial regression model $\hat{\beta}_w$ of $\mathbf{w}_i(x,y)$ as follows:
\begin{equation}
\begin{aligned}
    \hat{\beta}_w = (\mathbf{X}_{w}^{T}\mathbf{X}_{w})^{-1}\mathbf{X}_{w}^{T}\mathbf{y}_{w},
\end{aligned}
\label{eq_wall_find}
\end{equation}
where $\mathbf{X}_{w} = \{ x_i \in \mathbf{w}_i(x,y)\}_{i = 1:n}$ and $\mathbf{y}_{w} = \{ y_i \in \mathbf{w}_i(x,y)\}_{i = 1:n}$.
We next estimate the distance to the wall used for collision warning as follws:
\begin{equation}
\begin{aligned}
    d_w = \mathbf{y}_{w}(0.).
\end{aligned}
\label{eq_wall_distance}
\end{equation}
Thus, we can obtain a control command $\mathbf{u}_w$ using the derivative of the look ahead point $d_{l.h}$ and desired distance from wall $d_{gap}$ as follows:
\begin{equation}
\begin{aligned}
    \mathbf{u}_w = w_{\Theta}\frac{d}{dx}(\mathbf{y}_{w}(d_{l.h})) + w_{d}(d_w - d_{gap}),
\end{aligned}
\label{eq_wall_cmd}
\end{equation}
where weights $w_{\Theta}$ and $w_{d}$ account for the wall curvature and distance to the wall, respectively.

\subsection{Planning for obstacle avoidance mission}
\begin{figure}[!t]
    \centering
    \subfigure[]{
        \includegraphics[width=0.75\columnwidth]{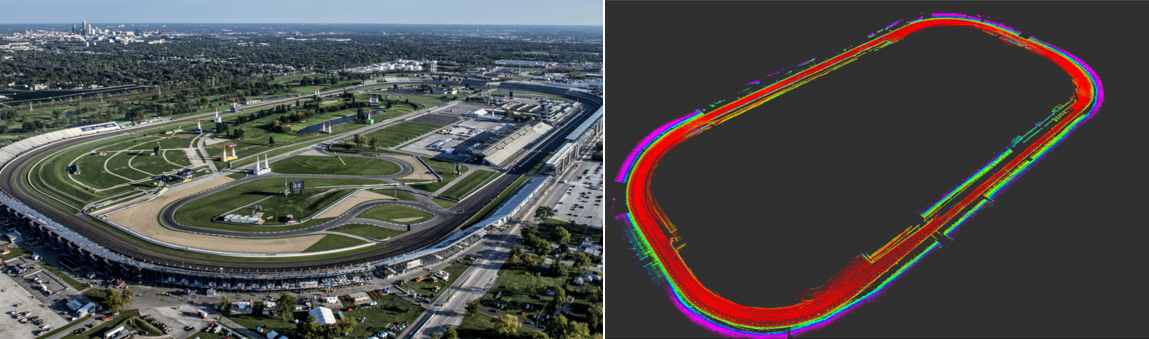}
        \label{fig:race_track}
    }
    \subfigure[]{
        \includegraphics[width=0.75\columnwidth]{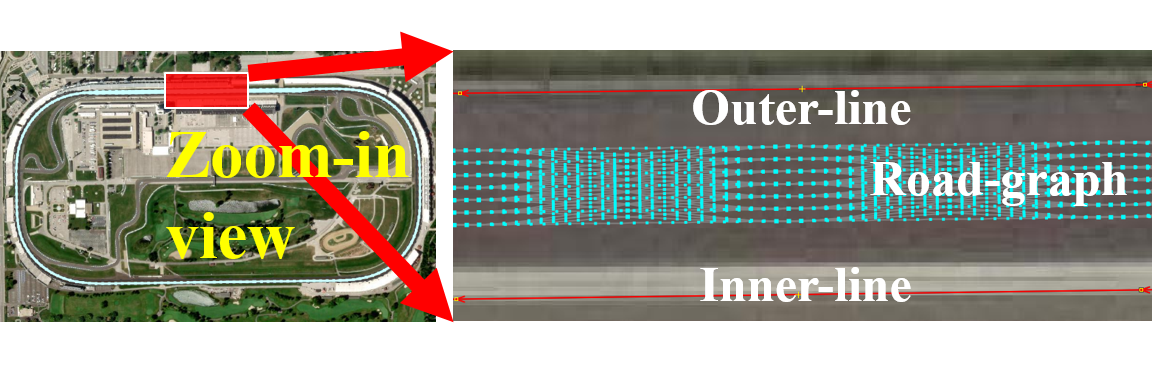}
        \label{fig:road_graph}
    }   
    \subfigure[]{
        \includegraphics[width=0.75\columnwidth]{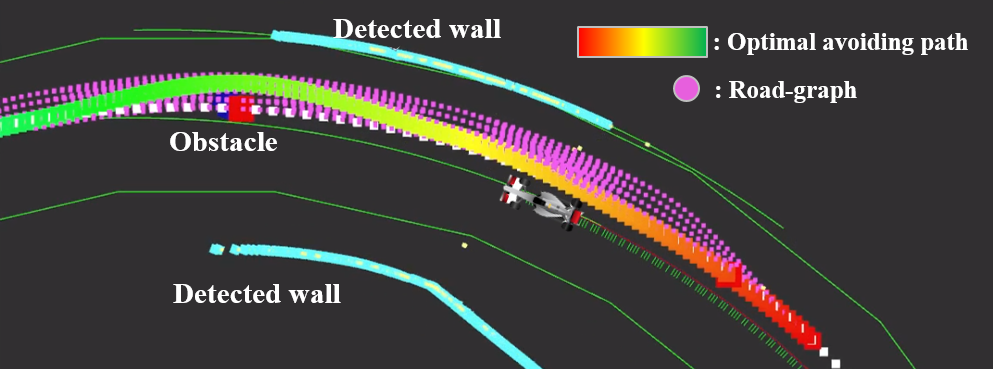}
        \label{fig:optimal_path}
    }       
    \label{fig:mapping_road_graph}
    \caption[Description of road-graph generation process]{
    Description of road-graph generation process. A 3D point-cloud map is constructed to build a road-graph for a path planning algorithm. 
    The race track boundary is found by extracting the coordinates from the 3D map. 
    From the road-graph model, an optimal route is computed.
    (a) IMS, Indianapolis, and its 3D point-cloud map.
    (b) Satellite view of IMS and zoom-in view of generated road graph.
    (c) Optimal path planning for obstacle avoidance challenge. The obstacle is depicted in red and the generated road graph in pink. 
    The optimal path is computed using the designed cost function described in rainbow color. 
    }
\end{figure}

The IAC organizers aimed to provide more challenging scenarios; thus, there was an obstacle avoidance mission to qualify for the semi-final event.
The vehicle was required to avoid obstacles placed along straight paths while maintaining high speeds of over 100 kph.
In addition, we had to guarantee that our race car drove under in-bounded conditions. 
Therefore, we propose a road-graph-based path planning algorithm to ensure that our race car meets these constraints. \\
Given the estimated position $\hat{\mathbf{x}}$, the race car is expected to follow the planned path $\mathbf{q}^{g}_{1:n}$.
To generate an optimal racing line $\mathbf{q}^{g}_{1:n}$ offline, we adopt studies \cite{heilmeier2019minimum, christ2021time, hermansdorfer2019concept, herrmann2020minimum}, which optimizes the racing line between the inner and outer lines.
We build a 3D point-cloud map $\{\mathbf{p}^M_{i}\}_{i \in [1:t]}$ according to $\{\hat{\mathbf{x}_i}\}_{i \in [1:t]}$ stacking point-cloud $\{\mathbf{p}^B_{i}\}_{i \in [1:t]}$.
\begin{equation}
\begin{aligned}
    \{\mathbf{p}^B_{i}\}_{i \in [1:t]} =\sum_{i}^t r(\mathbf{T}_{\hat{\mathbf{x}},i}\cdot\mathbf{p}^B_{i}),
\end{aligned}
\label{eq_map_optimize}
\end{equation}
where $\mathbf{T}_{\hat{\mathbf{x}},i}$ is a 4x4 transformation matrix derived from $\hat{\mathbf{x}}$, and $r(x)$ is the scan-matching registration algorithm \cite{koide2021voxelized} that corrects scan drift online.
We depict the constructed 3D point-cloud map in Fig. \ref{fig:race_track}.
Subsequently, we extract the inner and outer racing track lines using a 3D point-cloud map. \\
Furthermore, we generate a fully-connected road-graph offline as shown Fig. \ref{fig:race_track}. 
Let the offline generated road-graph $\{\mathbf{q}_{j}^{graph}\}_{j \in [1:m]} = \{N_{i,j}, E_{j}, C_{i, j}, \kappa_{i, j}, d_{i,j}\}_{j \in [1:m]}$ be defined by further enriching the connectivity $C_{i: j}$, curvature $\kappa_{i, j}$, and distance from racing line $d_{i,j}$, where $i$ is the interpolating index between multi-lanes.
Here, we use the same concept as in \cite{stahl2019multilayer} to precompute the state lattice offline, which is adopted in the Frenét space along a $\mathbf{q}^{g}_{1:n}$. The Frenét frame is defined as the coordinate system spanned by the tangential and normal vectors at any point of $\mathbf{q}^{g}_{1:n}$.
Let the state lattice be defined using discrete functions $[x(s), y(s),\theta(s),\kappa(s)]$ along the arc length $s$. 
Then, we can define $\mathbf{q}_{1:m}^{graph}$, with lattice layers distributed along the station $s$ connecting road-graph, as depicted in Fig. \ref{fig:road_graph}. \\

Let the optimal racing line computed from road-graph $\{\mathbf{q}_{j}^{graph}\}_{j \in [1:m]}$ be defined as $\mathbf{q}^{g*}_{1:n} = \{q_1^g, \dots, q_n^g\}$. 
When detecting obstacles from clusters $C_j \in \mathbf{p}^{filtered}_{t}$, we use an offline-computed set of costs represented as $\{\kappa_{i, j}, d_{i,j}\}$.
Moreover, we design a heuristic cost $g^{*}_{i,1:m} \in \{\mathbf{q}_{j}^{graph}\}_{j \in [1:m]}$ as   
\begin{equation}
\begin{aligned}
    g^{*}_{i, 1:m} = 
    & k_{{C}}\parallel {C}_{j} - {N}_{i, 1:m} \parallel_{2} + k_{\kappa}{\kappa}_{i, 1:m} + k_{d}d_{i, 1:m},
\end{aligned}
\label{eq_heuristic}
\end{equation}
where $k_{\kappa}$, $k_{trans}$, $k_{route}$ are the weights accounting for distance from closest node ${N}_{i, 1:m}$ to clusters $C_{j}$, road-curvature $\kappa_{i}$, and distance from racing line $d_{i,j}$, respectively. \\
Therefore, we define our collision-free motion planning algorithm considering optimal travel distance function $f^{*}_{i, 1:m}$ and designed heuristic cost $g^{*}_{i, 1:m}$ as  
\begin{equation}
\begin{aligned}
    \mathbf{q}^{*}_{t, 1:n} = \arg\min_{\forall{i}}{(f^{*}_{i, 1:m} + g^{*}_{i, 1:m})}.
\end{aligned}
\label{eq_optimal}
\end{equation}

\section{Experiments}
\label{sec:results}
\subsection{Vehicle platform}
Race car computation hardware consists of an ADLink x64 computer with eight Intel Xeon cores and an NVIDIA RTX 8000 GPU.
The sensor unit has two independent Novatel GPS units, three Luminar LIDARs, three Aptiv radars, and six Allied Vision cameras. 
In addition, vehicle states, such as wheel speeds and brake pressures, are available.
More detailed information is represented in Table \ref{tab:platform}.
\begin{table}[!t]
\caption[Platform's computing and networking hardware specifications]{\textbf{Computing and networking hardware specifications}}
\label{tab:platform}
\begin{center}
    \begin{tabular}{l|l}
    Device   & Specification \\
    \hline\hline
    CPU            & Intel Xeon E 2278 GE – 3.30 GHz (16T, 8C)                                              \\\hline
    GPU            & Nvidia Quadro RTX 8000 x 1 (PCIe slot)                                                 \\\hline
    RAM            & 64 GB                                                                                  \\\hline
    Ethernet ports & 3 x GigE RJ45 (In-built)                                                               \\
                   & + 2 x 40GbE QSFP+ (PCIe-based)                                                         \\\hline
    CAN ports      & 4 x In-built ports, 2 x port on PCIe card                                              \\\hline
    Network switch & Cisco IE-5000-12s12p-10G -- 12 GigE copper                                             \\
                   & 12 GigE fiber, 4x 10G uplink, PTP GM Clock                                             \\\hline
    Wireless       & Cisco/Fluidmesh FM4500 - up to 500Mbp \\ \hline
    GPS            & 2 x NovAtel PwrPak7 + HxGN SmartNet RTK \\ \hline
    LiDAR          & 3 x Luminar Hydra 120$^{\circ}$  \\ \hline
    RGB camera     & 6 x optical camera \\ \hline 
    Radar          & 3 x Delphi(1 x ESR + 2 x MRR) 
    \end{tabular}
    \end{center}
\end{table}

\subsection{Single racing time trial competition}
On October 23$\textsuperscript{rd}$, 2021, the semi-final and final of the time trial competition were held at IMS.
First, in the semi-final, each race car is required to run five laps comprising of: warming up (lap 1), performance laps (lap 2 and 3), cool-down (lap 4), and obstacle avoidance (lap 5).
Only the top three teams in the semi-finals were allowed to compete in the finals. The semi-finals were judged based on average lap speed and obstacle avoidance performance.
Full broadcast video is available here: \href{https://www.indyautonomouschallenge.com/full-broadcast-of-indy-autonomous-challenge-powered-by-cisco-event}{https://www.indyautonomouschallenge.com/}.
Moreover, our part of the race appears from 0:57:43 to 1:15:07 in the full broadcast video. 
In addition, the visualizations of our resilient navigation system and obstacle avoidance challenge are available here: \href{https://youtu.be/fiSqdMDmjGo}{https://youtu.be/fiSqdMDmjGo}.


\begin{figure}[!ht]
    \centering
    \subfigure[]{
        \includegraphics[width=1.0\columnwidth]{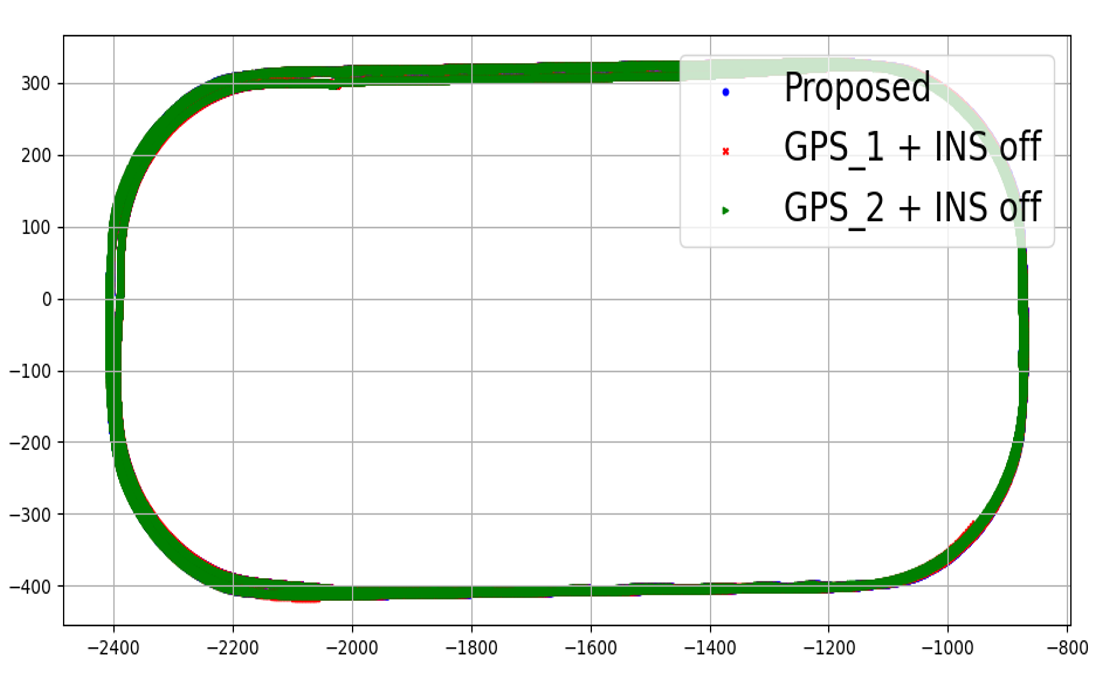}
        \label{fig:result_proposed}
    }
    \subfigure[]{
        \includegraphics[width=0.45\columnwidth]{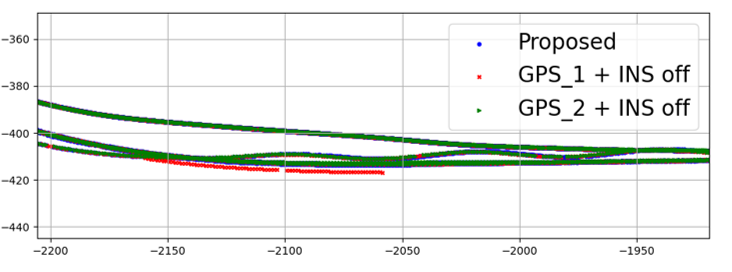}
        \label{fig:localize_sub_1}
    }
    \subfigure[]{
        \includegraphics[width=0.45\columnwidth]{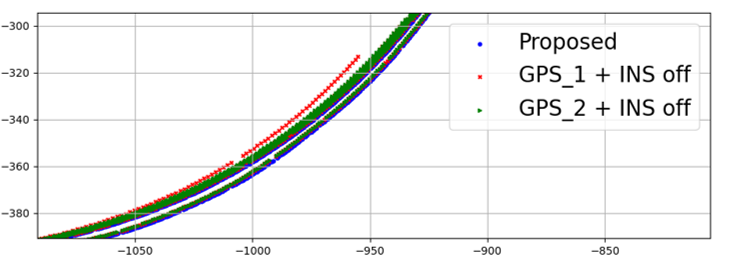}
        \label{fig:localize_sub_2}
    }
    \subfigure[]{
        \includegraphics[width=0.45\columnwidth]{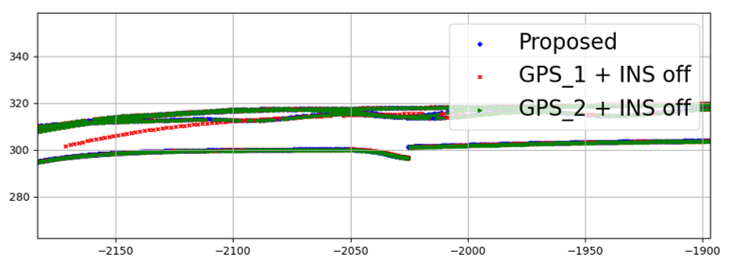}
        \label{fig:localize_sub_3}
    }    
    \subfigure[]{
        \includegraphics[width=0.45\columnwidth]{fig/result_localize_sub_3.png}
        \label{fig:localize_sub_4}
    }     
    \subfigure[]{
        \includegraphics[width=0.75\columnwidth]{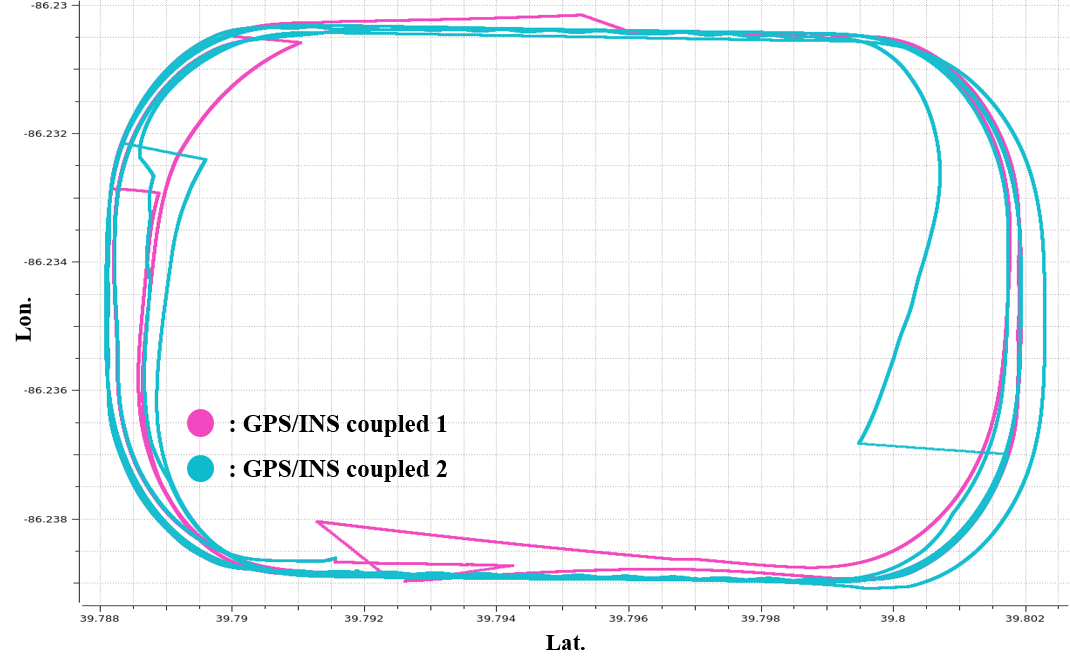}
        \label{fig:result_gps_ins_couple}
    }       
    
    \label{fig:result_localize}
    \caption[This should be changed to clear plot.]{
    Result of GPS log data and its estimated position using the proposed localization.
    (a) Received GPS data with partial degradation. Despite the degradation,  the proposed localization algorithm estimated position resiliently throughout the track.
    (b)-(e) Zoom-in view of GPS data and localization result.
    (f) GPS data based on GPS/INS tightly coupled method. In this study, we turn off the INS functionality because it is more prone to diverge with strong vibration.
    }
\end{figure}

\subsection{Multi-sensor fusion Kalman filter}
\begin{figure}[!t]
    \centering
    \subfigure[]{
        \includegraphics[width=1.0\columnwidth]{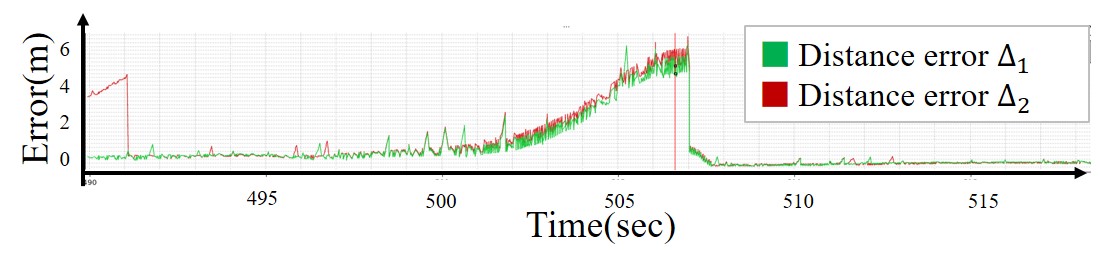}
        \label{fig:result_dist_err}
    }
    \subfigure[]{
        \includegraphics[width=1.0\columnwidth]{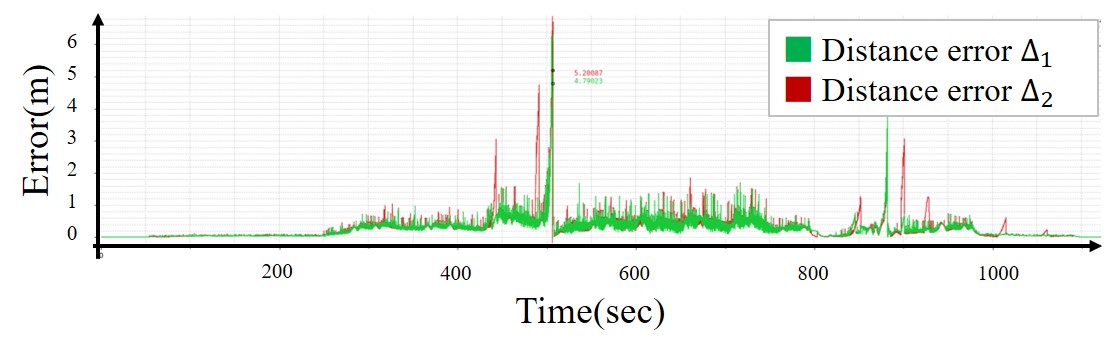}
        \label{fig:result_dist_err_zoom}
    }
    \subfigure[]{
        \includegraphics[width=1.0\columnwidth]{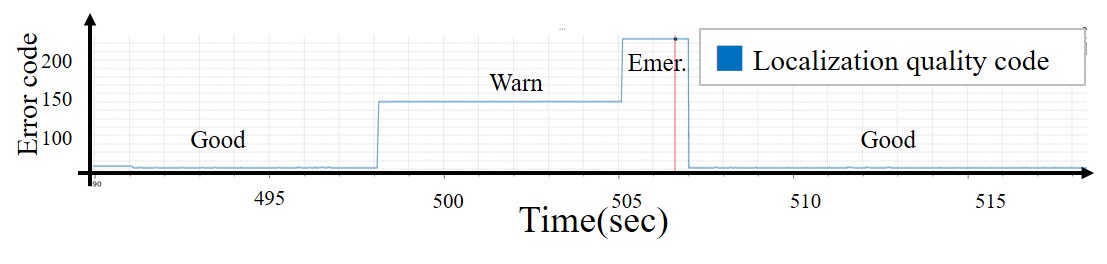}
        \label{fig:result_error_code}
    }    
    \label{fig:result_wall}
    \caption[This should be changed to clear plot.]{
    The proposed distance value increases rapidly during driving to estimate the GPS state.
    Both GPS data continuously diverged, and our localization system manager warned its status and prepared to change to a wall following maneuver.
    (a) Full sequential data of estimated localization hyper-parameter.
    (b) GPS data suddenly deteriorates while driving in Lap 2. (c) Localization system manager status. The status changes according to the localization hyper-parameter value.
    }
\end{figure}

As shown in Fig. \ref{fig:result_gps_ins_couple}, there was a number of positioning-degradation in both GPS units. Hence, we turned off the INS functionality, which is more prone to diverge with strong vibration.
However, we could estimate the vehicle's position with degraded data based on the proposed multi-sensor fusion Kalman filter in \ref{sec:kalman}.
During racing, we set $\epsilon$ as 0.2 $m$ and $\delta$ as 5.0 $m$ in Eq. \ref{eq_conditional_measure}. The localization result is shown in Fig. \ref{fig:result_proposed}.
There was critical degradation in both GPS units from timestamp 498 to 505 which can be seen in Fig. \ref{fig:result_dist_err_zoom}. Here our system manager warned us to get ready to drive with wall following mode resiliently.
We illustrate this warning and emergency status in Fig. \ref{fig:result_error_code}.

\subsection{Resilient navigation}
\begin{figure}[!t]
    \centering
    \subfigure[]{
        \includegraphics[width=0.45\columnwidth]{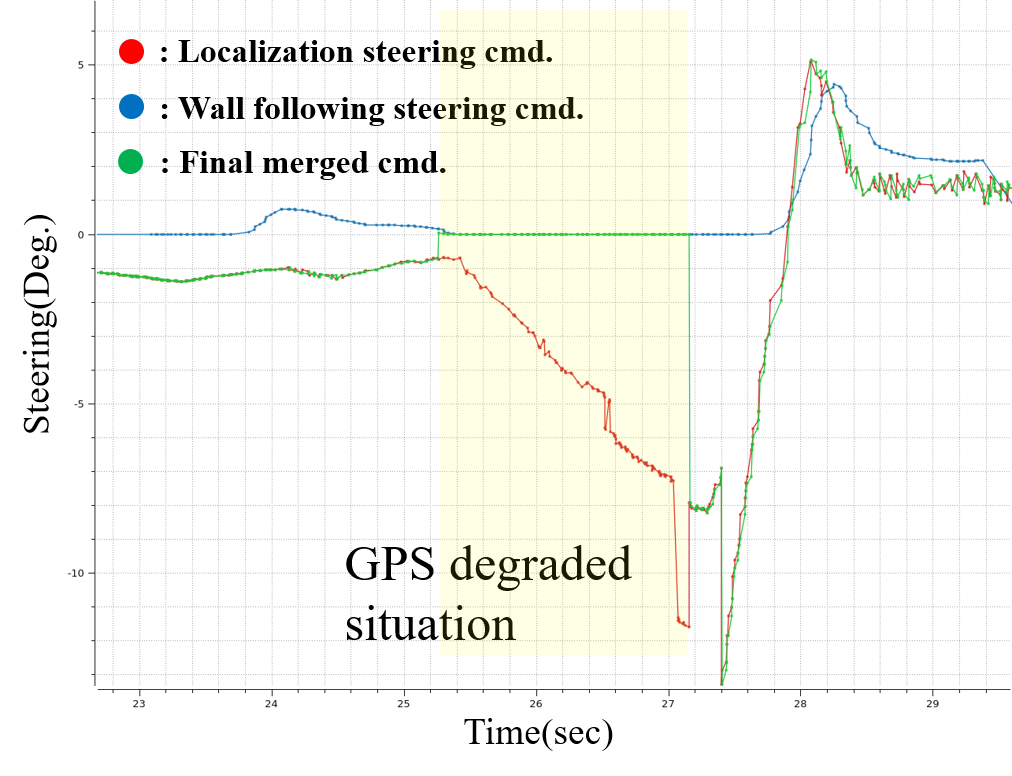}
        \label{fig:result_steering_1}
    }
    \subfigure[]{
        \includegraphics[width=0.45\columnwidth]{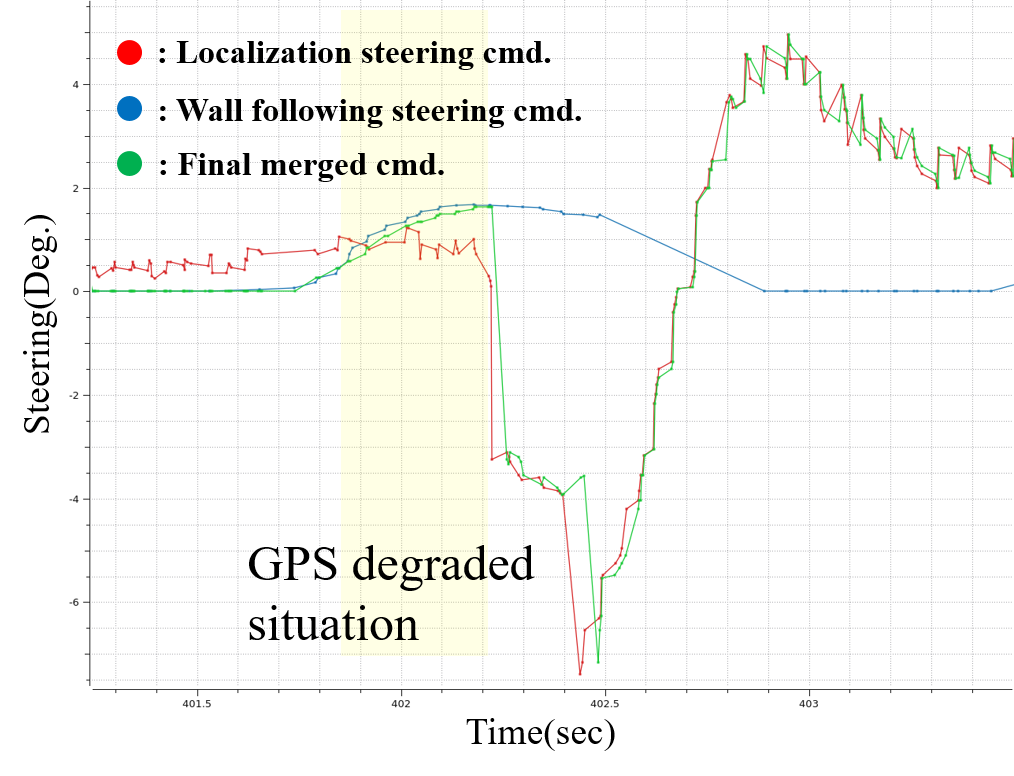}
        \label{fig:result_steering_2}
    }
    \label{fig:result_resilient_steer}
    \caption[This should be changed to clear plot.]{
    Proposed resilient navigation system which prevents crashes caused by critical GPS degradation. 
    (a) The proposed algorithm keeps the racing car at a certain distance from the wall for approximately 2 s to prevent crashing.
    (b) After receiving reasonable measurement data, the final steering command uses the localization-based command.
    }
\end{figure}

This study proposes a resilient navigation system using the detected walls of racing tracks.
Subsequently, we compute steering command $\mathbf{u}_w$ to follow the wall using Eq. \ref{eq_wall_cmd}. 
When our localization system manager's estimated localization status changes to emergency status, we designed our vehicle to drive with $\mathbf{u}_w$ to avoid crashing.
As shown in Fig. \ref{fig:result_resilient_steer}, our resilient navigation system came up with critical degradation on both GPS units.
Simultaneously, our resilient navigation algorithm enabled the race car to drive stably along the wall.

Moreover, when analyzing the lateral steering command that follows the racing line based on localization with degraded GPS measurement, we found the negative directional steering command causing the vehicle to crash onto the wall, as illustrated in \ref{fig:result_steering_1}.

If the race car had continued following the racing line with localization in a degraded state, the vehicle would have crashed. This is shown in Fig. \ref{fig:result_error_code}.
Analyzing the lateral steering commands that 

Steering commands immediately reverted back to GPS based localization once sensor degradation stopped. 

\begin{figure}[!ht]
    \centering
    \includegraphics[width=1.0\columnwidth]{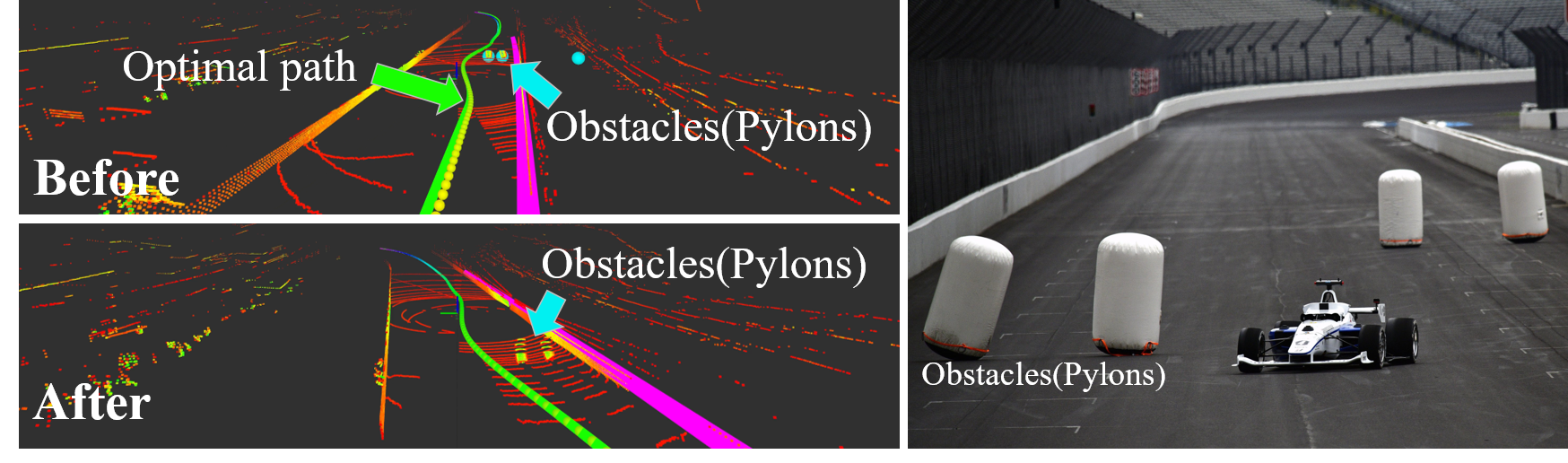}
    \label{fig:result_planning}
    \caption[Obstacle avoidance mission]{
    Photograph of the moment of obstacle avoidance. 
    Our race car successfully avoided the pylons at 111.4 $kph$.
    Our optimal path is shown in green. The obstacles and detected walls are illustrated in cyan and pink, respectively.
    }
\end{figure}

\subsection{Obstacle avoidance challenge}
The obstacle avoidance challenge required the vehicle to avoid two sets of two pylons placed on the track as illustrated in Fig. \ref{fig:result_planning}. 
One preventing an early lane change and another forcing a lane change.
The pylon avoidance had to be completed at a minimum speed of 60 $mph$.
Their locations were not disclosed until the start of the race. 
Our race car successfully avoided the pylons at 111.4 $kph$.
The generated path shown in the figure accounts for collision, curvature, and original racing line.

\subsection{Result}
We finished 4\textsuperscript{th} at the IAC main event as shown in Table \ref{tab.competition}. 
In addition, we were among the only four teams who completed the entire sequence in this event.
The numerous experiments needed for the study was conducted at Lucas oil raceway and IMS. Several distances were tested to validate our developed algorithm over the sensing, perception, planning, and control modules, as depicted in Table \ref{tab.field_summary}.

\begin{table}[ht]
\caption[IAC competition results summary.]{IAC competition results summary. \\ DNF: Did Not Finish; DNQ: Did Not Qualify.}
\label{tab.competition}
\begin{tabular}{lc}
\hline
\textbf{Team}                  & \textbf{IAC 2021, Indianapolis}\\ \hline
\multicolumn{1}{l|}{Technical University of Munich} & 1st\\
\multicolumn{1}{l|}{TII Euroracing}                 & 2nd\\
\multicolumn{1}{l|}{PoliMOVE}                       & 3rd\\
\multicolumn{1}{l|}{\textbf{KAIST}}                  & \textbf{4th}\\
\multicolumn{1}{l|}{University of Virginia}         & DNF\\
\multicolumn{1}{l|}{Auburn University}              & DNQ\\
\multicolumn{1}{l|}{MIT-PITT-RW}                    & DNF\\
\multicolumn{1}{l|}{University of Hawai'i}          & DNF\\
\multicolumn{1}{l|}{Purdue University}              & DNF
\end{tabular}
\end{table}

\begin{table}[ht]
\caption[Field tests performance summary.]{Field tests performance summary.}
\label{tab.field_summary}
\begin{adjustbox}{width=\columnwidth}
\begin{tabular}{lcccll}
\hline
\textbf{Task}                                               & \textbf{LOR}                                                    & \textbf{IMS}                                                  & \multicolumn{2}{l}{\textbf{Remarks}}    \\ \hline
\multicolumn{1}{l|}{{Pit-in / Pit-out}}                     & {\checkmark}                                                    & {\checkmark}                                                  & {Typical:}       & 15.00 m/s \\
\multicolumn{1}{l|}{}                                       &                                                                 &                                                               &                                       & 54.0 km/h  \\ \hline
\multicolumn{1}{l|}{{Performance}}                      & {\checkmark}                                                    & {\checkmark}                                                  & {Maximum:}       & 41.08 m/s \\
\multicolumn{1}{l|}{lap}                                       &                                                                 &                                                               &                                       & 147.9 km/h \\ \hline
\multicolumn{1}{l|}{{Obstacle}}                   & {\checkmark}                                                    & {\checkmark}                                                  & {Maximum:}       & 30.94 m/s \\
\multicolumn{1}{l|}{avoidance}                                       &                                                                 &                                                               &                                       & 111.4 km/h \\ \hline
\multicolumn{1}{l|}{Driving distance}              & 134.39                                                          & 174.45                                                        & Total:                                & $\sim$308.84 km        \\ 
\multicolumn{1}{l|}{{[}km{]}}                                       &                                                                 &                                                               &                                       &  \\ \hline
\multicolumn{1}{l|}{Operation time}                 & 4.10                                                            & 3.24                                                          & Total:                                & $\sim$7.3 h         \\ 
\multicolumn{1}{l|}{{[}h{]}}                                       &                                                                 &                                                               &                                       &  \\ \hline
\multicolumn{1}{l|}{Maximum speed}                          & \begin{tabular}[c]{@{}l@{}}27.74 m/s\\ 99.9 km/h\end{tabular}   & \begin{tabular}[c]{@{}l@{}}41.08 m/s\\ 147.9 km/h\end{tabular}
\end{tabular}
\end{adjustbox}
\end{table}

\section{Conclusion}
\label{sec:conclusion}
We presented a multi-sensor fusing Kalman filter for a full-scale autonomous race car to deal with degraded GPS measurement due to strong vibration. 
We also presented a resilient navigation system using wall detection.
Moreover, the vertical feature extraction algorithm for a LiDAR point cloud enhanced the real-time computation capabilities.
Lastly, we implemented a road-graph-based path planning algorithm for obstacle avoidance missions that allows driving within the in-bounded condition considering the original optimal racing line, obstacles, and vehicle dynamics.
We demonstrated that our designed localization system could handle degraded data and prevent a serious crash while driving at high-speed on competition day.
We successfully completed the obstacle avoidance challenge with a qualifying speed of 111.4 $kph$.

\section*{Acknowledgment}
We would like to thank Energy System Network (ESN), Juncos Hollinger Racing, and all other participating teams for their support and contributions to the event. 


\bibliographystyle{unsrt}
\bibliography{citation.bib}

\end{document}